\documentclass[11pt,a4paper]{article}
\usepackage[hyperref]{acl2020}
\usepackage{times}
\usepackage{import}
\usepackage{latexsym}
\usepackage{amsmath}
\usepackage{bbm}
\usepackage{url}
\usepackage{hyperref}
\usepackage{tabularx}
\usepackage{enumitem}
\usepackage[algoruled,vlined ]{algorithm2e}
\usepackage{mathtools}
\usepackage{multirow}
\usepackage{comment}
\usepackage{float}
\usepackage{amsfonts}

\DeclareMathOperator*{\avg}{avg} 
\DeclareMathOperator*{\argmax}{argmax} 
\DeclareMathOperator*{\argmin}{argmin} 

\usepackage{microtype}

\aclfinalcopy 

\newcommand{\wx}[1]{\textcolor{magenta}{\bf\small [#1 --WX]}}
\newcommand{\mm}[1]{\textcolor{orange}{\bf\small [#1 --MM]}}

\usepackage{array}
\newcolumntype{P}[1]{>{\centering\arraybackslash}p{#1}}

\title{Neural CRF Model for Sentence Alignment in Text Simplification}

\author{Chao Jiang, Mounica Maddela, Wuwei Lan, Yang Zhong, Wei Xu \\
Department of Computer Science and Engineering\\ 
The Ohio State University\\
  \texttt{\{jiang.1530,  maddela.4, lan.105,zhong.536, xu.1265\}@osu.edu}
}

\date{}

\begin{document}
\maketitle

\begin{abstract}
The success of a text simplification system heavily depends on the quality and quantity of complex-simple sentence pairs in the training corpus, which are extracted by aligning sentences between parallel articles. To evaluate and improve sentence alignment quality, we  create two manually annotated  sentence-aligned datasets from  two commonly used text simplification corpora, Newsela and Wikipedia. We propose a novel neural CRF alignment model which not only leverages the sequential nature of sentences in parallel documents but also utilizes a neural sentence pair model to capture semantic similarity.  Experiments demonstrate that our proposed approach outperforms all the previous work on monolingual sentence alignment task by  more than 5 points in F1.  We apply our CRF aligner to construct two new text simplification datasets, {\sc Newsela-Auto} and {\sc Wiki-Auto}, which are much larger and of better quality compared to the existing datasets.  A Transformer-based seq2seq model trained on our datasets establishes a new state-of-the-art for text simplification in both automatic and human evaluation.\footnote{Code and data are available at: \url{https://github.com/chaojiang06/wiki-auto}. Newsela data need to be requested at: \url{https://newsela.com/data/}.}

\end{abstract}

\section{Introduction}

Text simplification aims to rewrite complex text into simpler language while retaining its original meaning \cite{Saiggon}. Text simplification can provide reading assistance for children \cite{Kajiwara2013},  non-native speakers \cite{petersen2007text, pellow-eskenazi:2014:PITR}, non-expert readers \cite{Elhadad2007,siddharthan-katsos:2010:NAACLHLT},  and people with language disorders \cite{Rello:2013:ILS:2458308.2458354}. As a preprocessing step, text simplification can also improve the performance of many natural language processing (NLP) tasks, such as parsing \cite{chandrasekar-etal-1996-motivations}, semantic role labelling \cite{vickrey-koller-2008-sentence}, information extraction \cite{miwa-etal-2010-entity} , summarization \cite{Vanderwende-InfProcessManage-2007-BST, xu-grishman-2009-parse}, and machine translation \cite{chen-etal-2012-simplification, stajner-popovic-2016-text}.

Automatic text simplification is primarily addressed by sequence-to-sequence (seq2seq)  models whose success largely depends on the quality and quantity of the training corpus, which consists of pairs of complex-simple sentences. Two widely used corpora, {\sc Newsela} \cite{Xu-EtAl:2015:TACL} and {\sc WikiLarge} \cite{zhang-lapata-2017-sentence}, were created by automatically aligning sentences between comparable articles. However, due to the lack of reliable annotated data,\footnote{\newcite{hwang-EtAl:2015:NAACL-HLT} annotated 46 article pairs from Simple-Normal Wikipedia corpus; however, its annotation is noisy, and it contains many sentence splitting errors.}  sentence pairs are often aligned using  surface-level similarity metrics, such as Jaccard coefficient \cite{Xu-EtAl:2015:TACL} or cosine distance of TF-IDF vectors \cite{paetzold-etal-2017-massalign}, which fails to capture paraphrases and the context of surrounding sentences. A common drawback of text simplification models trained on such datasets is that they behave conservatively, performing mostly  deletion, and rarely paraphrase \cite{alva-manchego-etal-2017-learning}.  Moreover, {\sc WikiLarge} is the concatenation of  three early datasets \cite{zhu-etal-2010-monolingual,woodsend-lapata-2011-learning, coster-kauchak-2011-simple} that are extracted from Wikipedia dumps and are known to contain many errors \cite{Xu-EtAl:2015:TACL}.

\begin{figure*}[t!]
\centering
\includegraphics[width=1.0\textwidth]{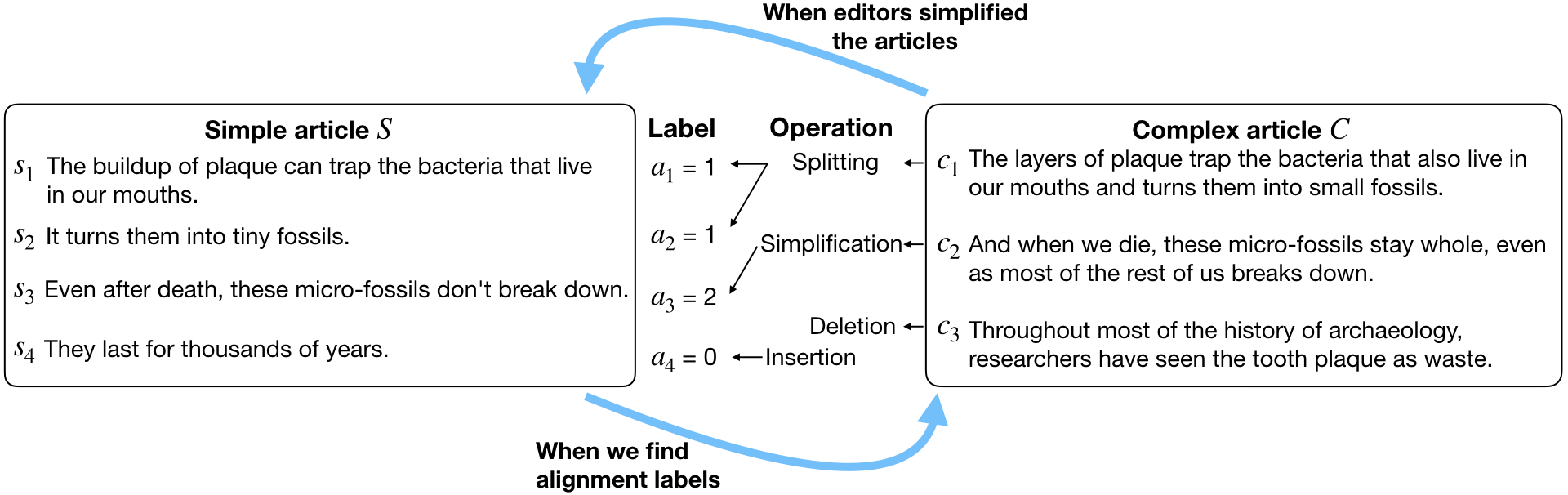}
\caption{An example of sentence alignment between an original news article (right) and its simplified version (left) in Newsela. The label $a_i$ for each simple sentence $s_i$ is the index of complex sentence $c_{a_i}$ it aligns to.}
\label{fig:pairexample}
\end{figure*}

To address these problems, we create the first high-quality manually annotated sentence-aligned datasets: {\sc Newsela-Manual} with 50 article sets, and {\sc Wiki-Manual} with 500 article pairs. We design a novel neural CRF alignment model, which utilizes fine-tuned BERT to measure semantic similarity  and  leverages the similar order of content between parallel documents, combined with an effective paragraph alignment algorithm. Experiments show that our proposed method outperforms all the previous  monolingual sentence alignment approaches \cite{STAJNER18.630, paetzold-etal-2017-massalign, Xu-EtAl:2015:TACL} by more than 5 points in F1.

By applying our alignment model to all the 1,882 article sets in Newsela and 138,095 article pairs in  Wikipedia dump, we then construct two new simplification datasets, {\sc Newsela-Auto} (666,645 sentence pairs) and {\sc Wiki-Auto} (488,332 sentence pairs). Our new datasets with improved quantity and quality facilitate the training of complex seq2seq models. A BERT-initialized Transformer model trained on our datasets outperforms the state-of-the-art by 3.4\% in terms of SARI, the main automatic metric for text simplification. Our simplification model produces 25\% more rephrasing than those trained on the existing datasets. Our contributions include:

\begin{enumerate}[topsep=2pt,itemsep=0pt,partopsep=2pt,parsep=1pt]
    
    \item Two manually annotated datasets that enable the first systematic study for training and evaluating monolingual sentence alignment;

    \item A neural CRF sentence alinger and a paragraph alignment algorithm that employ fine-tuned BERT to capture semantic similarity and take advantage of the sequential nature of parallel documents;
    
    \item Two automatically constructed  text simplification datasets which are of higher quality and 4.7 and 1.6 times larger than the existing datasets in their respective domains;
    
    \item A BERT-initialized Transformer model for automatic text simplification, trained on our datasets, which establishes a new state-of-the-art in both automatic and human evaluation.

\end{enumerate}

\section{Neural CRF Sentence Aligner}
\label{sec:alignment_model}
We propose a  neural CRF sentence alignment model, which leverages the similar  order of content  presented in parallel documents and captures  editing operations across multiple sentences, such as splitting and elaboration (see Figure \ref{fig:pairexample} for an example).  To further improve the accuracy, we first align paragraphs based on semantic similarity and vicinity information, and then extract sentence pairs from these aligned paragraphs.  In this section,  we describe the task setup and our approach.

\subsection{Problem Formulation}

Given a simple article (or paragraph) $S$ of $m$ sentences and a complex article  (or paragraph) $C$ of $n$ sentences, for each sentence $s_i$ ($i\in [1,m]$) in the simple article, we aim to find its corresponding sentence $c_{a_i}$ ($a_i \in [0,n]$) in the complex article. We use $a_i$ to denote the index of the aligned sentence, where $a_i = 0$ indicates that sentence $s_i$ is not aligned to any sentence in the complex article. The full alignment $\mathbf{a}$ between article (or paragraph) pair  $S$ and $C$ can then be represented by a sequence of alignment labels $\mathbf{a}=(a_1, a_2, \dots, a_m)$. Figure \ref{fig:pairexample} shows an example of  alignment labels. One specific aspect of our CRF model is that it uses a varied number of labels for each article (or paragraph) pair rather than a fixed set of labels.

\subsection{Neural CRF Sentence Alignment Model}
\label{subsec:CRF}

We learn $P(\mathbf{a}|S,C)$, the conditional probability of alignment $\mathbf{a}$ given an article pair $(S, C)$, using linear-chain conditional random field:
\begin{equation} \label{eq1}
\begin{split}
P(\mathbf{a}|S,C)  & = \frac{\exp(\Psi(\mathbf{a},S,C))}{\sum_{\mathbf{a}\in \mathcal{A}}\exp(\Psi(\mathbf{a},S,C))} \\
 & = \frac{\exp(\sum _{i=1}^{|S|}\psi (a_i, a_{i-1}, S,C))}{\sum_{a\in \mathcal{A}}\exp(\sum _{i=1}^{|S|} \psi (a_i, a_{i-1}, S,C)))} 
\end{split}
\end{equation}
where $|S|=m$ denotes the number of sentences in article $S$. The score $\sum _{i=1}^{|S|} \psi (a_i, a_{i-1}, S,C)$ sums over the sequence of alignment labels $\mathbf{a}=(a_1, a_2, \dots, a_m)$ between the simple article $S$ and the complex article $C$, and could be decomposed into two factors as follows: \begin{equation}\label{eq2}
\begin{split}
\psi(a_i, a_{i-1}, S,C)  =  sim(s_i, c_{a_i}) + T(a_i, a_{i-1})
\end{split}
\end{equation}
where $sim(s_i, c_{a_i})$ is the \textbf{semantic similarity} score between the two sentences, and $T(a_i, a_{i-1})$ is a pairwise score for \textbf{alignment label transition} that $a_i$ follows $a_{i-1}$.

\paragraph{Semantic Similarity} A  fundamental problem in sentence alignment is to measure the semantic similarity between two sentences $s_i$ and $c_j$. Prior work used lexical similarity measures, such as Jaccard similarity \cite{Xu-EtAl:2015:TACL}, TF-IDF \cite{paetzold-etal-2017-massalign}, and continuous n-gram features \cite{STAJNER18.630}. In this paper, we fine-tune BERT \cite{devlin2018bert} on our manually labeled dataset (details in \S \ref{sec:data}) to capture semantic similarity.

\paragraph{Alignment Label Transition} In parallel documents, the contents of the articles are often presented in a similar order.  The complex sentence $c_{a_i}$ that is aligned to $s_i$,  is often related to the complex sentences $c_{a_{i-1}}$  and  $c_{a_{i+1}}$, which are aligned to $s_{i-1}$ and $s_{i+1}$, respectively. To incorporate this intuition, we propose a scoring function to model the transition between alignment labels using  the following features:

\begin{equation} 
\begin{split}
g_1& =  |a_i - a_{i-1}| \\ 
g_2 &= \mathbbm{1}(a_i = 0 ,  a_{i-1} \neq 0) \\
g_3 &= \mathbbm{1}(a_i \neq 0 ,  a_{i-1} = 0) \\
g_4 &= \mathbbm{1}(a_i = 0 ,  a_{i-1} = 0)
\end{split}
\end{equation} 
\noindent where $g_1$ is the absolute distance between $a_i$ and $a_{i-1}$, $g_2$ and $g_3$ denote if the current or prior  sentence is not aligned to any sentence, and $g_4$ indicates whether  both $s_i$ and $s_{i-1}$ are not aligned to any sentences. The score is computed as follows:
\begin{equation} \label{eq8}
\begin{split}
T(a_i, a_{i-1}) = \text{FFNN}([g_1, g_2, g_3, g_4])
\end{split}
\end{equation}
where $[,]$ represents concatenation operation and $\text{FFNN}$ is a 2-layer feedforward neural network. We provide more implementation details of the model in Appendix \ref{appendix:details_of_alignment}.

\subsection{Inference and Learning}

During inference, we find the optimal alignment $\hat{\mathbf{a}}$:
\begin{equation} \label{eq8}
\begin{split}
\hat{\mathbf{a}} = \operatorname*{argmax}_\mathbf{a} P(\mathbf{a}|S, C) 
\end{split}
\end{equation}
\noindent using Viterbi algorithm in  $\mathcal{O} (mn^2)$ time. During training, we maximize the  conditional probability of the gold alignment label $\mathbf{a}^*$:
\begin{equation} \label{eq9}
\begin{split}
\log P(\mathbf{a}^*|S,C) = &\Psi(\mathbf{a}^*, S,  C) - \\ & \log \sum_{\mathbf{a}\in \mathcal{A}}\exp(\Psi(\mathbf{a},S,C))
\end{split}
\end{equation}
The second term sums the scores of all possible alignments and can be computed using forward algorithm in $\mathcal{O} (mn^2)$ time as well.

\subsection{Paragraph Alignment}

Both accuracy and computing efficiency can be improved if we align paragraphs before aligning sentences. In fact, our empirical analysis revealed that sentence-level alignments mostly reside within the corresponding aligned paragraphs (details in \S \ref{sec:analysis_on_paragraph_alignment} and Table \ref{table:dev_perfprmance}).  Moreover,  aligning  paragraphs first provides more training instances and reduces the label space for our neural CRF model.  

We propose Algorithm \ref{alg:paragraph_similarity} and \ref{alg:paragraph_alignment} for paragraph alignment. Given a simple article $S$ with $k$ paragraphs $S = (S_1, S_2, \dots, S_k)$ and a complex article $C$ with $l$ paragraphs $C = (C_1, C_2, \dots, C_l)$, we first apply Algorithm \ref{alg:paragraph_similarity} to calculate the semantic similarity matrix $simP$  between paragraphs by averaging or maximizing over the sentence-level similarities (\S \ref{subsec:CRF}). Then, we use Algorithm \ref{alg:paragraph_alignment} to generate the  paragraph alignment matrix $alignP$. We align paragraph pairs if they satisfy one of the two conditions: (a) having high semantic similarity and appearing in similar positions in the article pair (e.g., both at the beginning), or (b) two continuous paragraphs in the complex article having  relatively high semantic similarity with one paragraph in the simple side, (e.g., paragraph splitting or fusion). The  difference of relative position in documents is defined as $d(i,j)=|\frac{i}{k}-\frac{j}{l}|$, and the thresholds $\tau_1$ - $\tau_5$ in Algorithm \ref{alg:paragraph_alignment} are selected using the dev set. Finally, we merge the neighbouring paragraphs which are aligned to the same paragraph in the simple article before feeding them into our neural CRF aligner. We provide more details in Appendix \ref{appendix:details_of_alignment}.

\begin{algorithm}[t!]
\small
\SetKwInput{kwInit}{Initialize}

\SetAlgoLined
\kwInit{$simP$ $\in$ $\mathbb{R}^{2 \times k \times l}$ to $0^{2 \times k \times l}$}
\For{$i\leftarrow 1$ \KwTo $k$}
{
    \For{$j\leftarrow 1$ \KwTo $l$}
    {
        \resizebox{1.0\hsize}{!}{
            $ simP[1, i, j] =   \avg\limits_{s_p \in S_i}\Big( \max\limits_{c_q \in C_j} simSent (s_p, c_q) \Big)$ \nonumber
        }

        \resizebox{1.0\hsize}{!}{
            $ simP[2, i, j] = \max\limits_{s_p \in S_i, c_q \in C_j} simSent(s_p, c_q)$ \nonumber
        }
    }

}
\Return  $simP$
\caption{Pairwise Paragraph Similarity}
\label{alg:paragraph_similarity}
\end{algorithm}

\vspace{-3pt}

\begin{algorithm}[t!]
\small
\SetKwInOut{Input}{Input}
\SetKwInput{kwInit}{Initialize}

\SetAlgoLined
\Input{$simP$ $\in$ $\mathbb{R}^{2\times k \times l}$}
\kwInit{$alignP$ $\in$ $\mathbb{I}^{k \times l}$ to $0^{k \times l}$}
\For{$i\leftarrow 1$ \KwTo $k$}
{
    $ j_{max} = \argmax\limits_{j}  simP[1, i,j] $ \\
    \If{$simP[1, i,j_{max}] > \tau_1$ and $d(i, j_{max}) < \tau_2$ }{$alignP[i, j_{max}]=1$}
    
    \For{$j\leftarrow 1$ \KwTo $l$}{
        \If{$simP[2, i,j] > \tau_3$ } {$alignP[i, j]=1$}
        \If{$j>1$ \& $simP[2, i,j] > \tau_4$ \& $simP[2, i,j-1] > \tau_4$ \& $d(i, j) < \tau_5$ \& $d(i, j-1) < \tau_5$}{$alignP[i, j]=1$\\ $alignP[i, j-1]=1$}
    }
}
\Return  $alignP$
\caption{Paragraph Alignment Algorithm}

\label{alg:paragraph_alignment}
\end{algorithm}

\vspace{-4pt}
\section{Constructing Alignment Datasets}
\label{sec:data}

To address the lack of reliable sentence alignment for Newsela \cite{Xu-EtAl:2015:TACL} and Wikipedia \cite{zhu-etal-2010-monolingual,woodsend-lapata-2011-learning}, we  designed an efficient annotation methodology to first manually align sentences between a few complex and simple article pairs. Then, we automatically aligned the rest using our alignment model trained on the human annotated data. We created two sentence-aligned parallel corpora (details in \S \ref{sec:experiment_on_ts}), which are the largest to date for text simplification.

\vspace{-4pt}

\subsection{Sentence Aligned Newsela Corpus}
\label{subsec:newsela_data_collection}

\begin{table}[t!]
\centering
\small
\begin{tabular}{l|c|c}
\hline
            & \textbf{Newsela} & \textbf{Newsela} \\
            & \textbf{-Manual} & \textbf{-Auto} \\ \hline
\multicolumn{3}{l}{\footnotesize{\textit{\textbf{Article level}}}} \\ \hline
\# of original articles & 50 & 1,882 \\
\# of article pairs &    500   &    18,820     \\  

\hline

\multicolumn{3}{l}{\footnotesize{\textit{\textbf{Sentence level}}}}    \\  \hline
\# of original sent. (level 0) & 2,190  & 59,752 \\ 
\# of sentence pairs &   1.01M$^\dagger$   &    666,645  \\
\# of unique complex sent.  & 7,001   &  195,566    \\ 
\# of unique simple sent.  & 8,008  &  246,420    \\
avg. length of simple sent.  &    13.9   &    14.8     \\  
avg. length of complex sent. &    21.3   &    24.9     \\  

\hline
\multicolumn{3}{l}{\footnotesize{\textit{\textbf{Labels of sentence pairs}}}} \\
\hline
\# of \textit{aligned} (not identical)   & 5,182  &\multirow{2}{*}{666,645}  \\

\# of \textit{partially-aligned} & 14,023  &  \\ 

\# of \textit{not-aligned} & 0.99M & -- \\ 
\hline
\multicolumn{3}{l}{\footnotesize{\textit{\textbf{Text simplification phenomenon}}}} \\
\hline
\# of sent. rephrasing (1-to-1)  &    8,216   &     307,450  \\
\# of sent. copying (1-to-1)  & 3,842 &   147,327    \\
\# of sent. splitting (1-to-n) &    4,237   & 160,300       \\
\# of sent. merging (n-to-1)  &   232     &   --     \\
\# of sent. fusion (m-to-n) & 252 & -- \\ 
\# of sent. deletion (1-to-0)  &  6,247   & --       \\
\hline
\end{tabular}
\caption{Statistics of our manually and automatically created sentence alignment annotations on Newsela. $\dagger$ This number includes all complex-simple sentence pairs (including \textit{aligned}, \textit{partially-aligned}, or \textit{not-aligned}) across all 10 combinations of 5 readability levels (level 0-4), of which 20,343 sentence pairs between adjacent readability levels were manually annotated and the rest of labels were derived.}
\vspace{-4pt}
\label{table:newsela_two_parts}
\end{table}

Newsela corpus \cite{Xu-EtAl:2015:TACL} consists of 1,932 English news articles where each article (level 0) is re-written by professional editors into four simpler versions at different readability levels (level 1-4).  We annotate sentence alignments for article pairs at adjacent readability levels (e.g., 0-1, 1-2) as the alignments between non-adjacent levels (e.g., 0-2) can be then derived automatically. To ensure efficiency and quality, we designed the following three-step annotation procedure:

\begin{figure}[t!]

\centering
\includegraphics[width=0.5\textwidth]{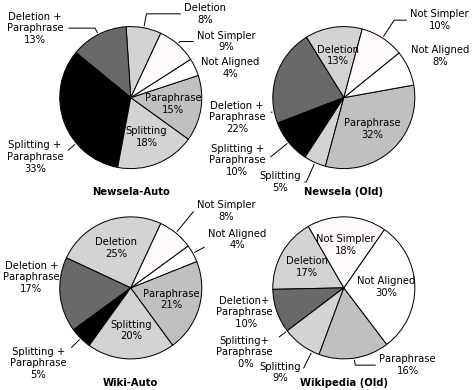}
\setlength{\belowcaptionskip}{-14pt}
\caption{Manual inspection of 100 random sentence pairs from our corpora ({\sc Newsela-Auto} and {\sc Wiki-Auto}) and the existing Newsela \cite{Xu-EtAl:2015:TACL} and Wikipedia \cite{zhang-lapata-2017-sentence} corpora. Our corpora contain at least 44\% more complex rewrites (\textit{Deletion + Paraphrase} or \textit{Splitting + Paraphrase}) and  27\% less defective pairs (\textit{Not Aligned} or \textit{Not Simpler}). }
\label{fig:dataset_operation_stat}

\end{figure}

\begin{enumerate}[topsep=2pt,itemsep=0pt,partopsep=2pt,parsep=1pt]
    \item Align paragraphs using CATS toolkit \cite{STAJNER18.630}, and then correct the automatic paragraph alignment errors by two in-house annotators.\footnote{We consider any sentence pair not in the aligned paragraph pairs as \textit{not-aligned}. This assumption leads to a small number of missing sentence alignments, which are manually corrected in Step 3.} Performing paragraph alignment as the first step significantly reduces the number of sentence pairs to be annotated from every possible sentence pair to the ones within the aligned paragraphs.  We design an  efficient visualization toolkit for this step, for which a screenshot can be found in Appendix \ref{appendix:human-annotation-inhourse-alignment}.
    
    \item For each sentence pair within the aligned paragraphs, we ask five annotators on the Figure Eight\footnote{https://www.figure-eight.com/} crowdsourcing platform to classify into one of the three categories: \textit{aligned}, \textit{partially-aligned}, or \textit{not-aligned}. We provide the annotation instructions and interface in Appendix \ref{appendix:human-annotation}. We require annotators to spend at least ten seconds per question and embed one test question in every five questions. Any worker whose accuracy drops below 85\% on test questions is removed. The inter-annotator agreement is 0.807 measured by Cohen's kappa \cite{artstein_inter-coder_2008}.
    \item We have four in-house annotators (not authors) verify the crowdsourced labels. 
    
\end{enumerate}

We  manually aligned 50 article groups to create the {\sc Newsela-Manual} dataset with a 35/5/10 split for train/dev/test, respectively.  We trained  our aligner on this dataset (details in \S\ref{sec:exp_sent_align}), then automatically aligned sentences in the remaining 1,882 article groups in Newsela (Table \ref{table:newsela_two_parts}) to create a new sentence-aligned dataset, {\sc Newsela-Auto}, which consists of 666k sentence pairs predicted as \textit{aligned} and \textit{partially-aligned}. {\sc Newsela-Auto} is considerably larger than the previous {\sc Newsela} \cite{Xu-EtAl:2015:TACL} dataset of 141,582 pairs, and contains 44\% more interesting rewrites (i.e., rephrasing and splitting cases) as shown in Figure \ref{fig:dataset_operation_stat}.

\label{subsec:newsela_data}

\renewcommand{\tabcolsep}{4pt}

\begin{table*}[ht!]
\small
\centering
\begin{tabular}{l|P{14mm}P{14mm}P{14mm}|P{14mm}P{14mm}P{14mm}}

\hline
\multirow{2}{*}{} & \multicolumn{3}{c|} {\textbf{Task 1} (\textit{aligned\&partial} vs. \textit{others})}                        

& \multicolumn{3}{c} {\textbf{Task 2} (\textit{aligned} vs. \textit{others})} \\

 & \textbf{Precision}  & \textbf{Recall} & \textbf{F1} &  \textbf{Precision}  & \textbf{Recall} & \textbf{F1} \\ \hline

\multicolumn{7}{l}{\textit{\textbf{Similarity-based models}}}    \\  \hline

Jaccard  \cite{Xu-EtAl:2015:TACL}         &               94.93    &         76.69              &     84.84          &   73.43              &          75.61            &              74.51                 \\ 

TF-IDF     \cite{paetzold-etal-2017-massalign}                 &           96.24          &               83.05        &            89.16     &         66.78          &     69.69                 &          68.20         \\

LR \cite{STAJNER18.630}                  &       93.11             &      84.96              &  88.85   &        73.21             &   74.74                  &         73.97                  \\ 
\hline 

\multicolumn{7}{l}{\textit{\textbf{Similarity-based models w/ alignment strategy (previous SOTA)} }}    \\  \hline

JaccardAlign \cite{Xu-EtAl:2015:TACL}                    &      98.66
                 &          67.58            &    80.22$^\dagger$                    &    51.34                  &      86.76                &    64.51$^\dagger$                  \\ 

MASSAlign \cite{paetzold-etal-2017-massalign}                            &        95.49              &   82.27                   &                  88.39$^\dagger$       &    40.98                  &  87.11                  &    55.74$^\dagger$   \\

CATS \cite{STAJNER18.630}                        &   88.56                    &                 91.31     &           89.92$^\dagger$  &     38.29                 &       97.39                &      54.97$^\dagger$            \\

\hline
Our CRF Aligner    &      \textbf{97.86}    & \textbf{93.43}     &     \textbf{95.59} &      \textbf{87.56}      &         \textbf{89.55}      &           \textbf{88.54}       \\ 

\hline

\end{tabular}
\setlength{\belowcaptionskip}{-8pt}
\caption{Performance of different sentence alignment methods on the {\sc Newsela-Manual} test set. $\dagger$ Previous work was designed only for Task 1 and used alignment strategy (greedy algorithm or dynamic programming) to improve either precision or recall.} 
\label{table:result_on_testset}
\end{table*}

\subsection{Sentence Aligned Wikipedia Corpus}
\label{subsec:wiki_data_collection}

We also create a new version of Wikipedia corpus by aligning sentences between English Wikipedia and Simple English Wikipedia. Previous work \cite{Xu-EtAl:2015:TACL} has shown that Wikipedia is much noisier than the Newsela corpus. We provide this dataset in addition to facilitate future research.

We first extract article pairs from English and Simple English Wikipedia by leveraging Wikidata, a well-maintained database that indexes named entities (and events etc.) and their  Wikipedia pages in different languages. We found this method to be more reliable than using page titles \cite{coster-kauchak-2011-simple} or cross-lingual links \cite{zhu-etal-2010-monolingual,woodsend-lapata-2011-learning}, as titles can be ambiguous and cross-lingual links may direct to a disambiguation or mismatched page (more details in Appendix \ref{appendix:details_of_processing_wikipedia}). In total,  we extracted 138,095 article pairs from  the 2019/09 Wikipedia dump, which is two times larger than the previous datasets \cite{coster-kauchak-2011-simple,zhu-etal-2010-monolingual} of only 60$\sim$65k article pairs, using an improved version of the WikiExtractor library.\footnote{https://github.com/attardi/wikiextractor}

Then, we crowdsourced the sentence alignment annotations for 500 randomly sampled document pairs (10,123 sentence pairs total). As document length in English and Simple English Wikipedia articles vary greatly,\footnote{The average number of sentences in an article is 9.2 $\pm$ 16.5 for Simple English Wikipedia and 74.8 $\pm$ 94.4 for English Wikipedia.} we designed the following annotation strategy that is slightly different from Newsela. For each sentence in the simple article, we select the sentences with the highest similarity scores from the complex article for manual annotation, based on four similarity measures: lexical similarity from CATS \cite{STAJNER18.630}, cosine similarity using  TF-IDF \cite{paetzold-etal-2017-massalign}, cosine similarity between BERT sentence embeddings, and alignment probability by a BERT model fine-tuned on our {\sc Newsela-Manual} data (\S \ref{subsec:newsela_data_collection}). As these four metrics may rank the same sentence at the top, on an average, we collected 2.13 complex sentences for every simple sentence and annotated the alignment label for each sentence pair. Our pilot study showed that this method captured 93.6\% of the aligned sentence pairs. We named this manually labeled dataset {\sc Wiki-Manual} with a train/dev/test split of 350/50/100 article pairs.

Finally, we trained our alignment model on this annotated dataset to automatically align sentences for  all the 138,095 document pairs (details in Appendix \ref{appendix:details_of_processing_wikipedia}). In total, we yielded 604k non-identical \textit{aligned} and \textit{partially-aligned} sentence pairs to create the {\sc Wiki-Auto} dataset. Figure \ref{fig:dataset_operation_stat} illustrates that {\sc Wiki-Auto} contains 75\% less defective sentence pairs than the old {\sc WikiLarge} \cite{zhang-lapata-2017-sentence} dataset.

\section{Evaluation of Sentence Alignment}
\label{sec:exp_sent_align}

\begin{table}[t!]
\small
\centering
\begin{tabular}{p{2.24cm}|ccc|ccc}
\hline
\multirow{2}{*}{} & \multicolumn{3}{c|}{\textbf{Task 1}} & \multicolumn{3}{c}{\textbf{Task 2}}  \\
 & \multicolumn{1}{c}{\textbf{P}}  & \multicolumn{1}{c}{\textbf{R}} &\multicolumn{1}{c|}{\textbf{F1}}   
 & \multicolumn{1}{c}{\textbf{P}}  & \multicolumn{1}{c}{\textbf{R}} &\multicolumn{1}{c}{\textbf{F1}} \\ 
 
 \hline

\multicolumn{7}{l}{\textbf{\textit{Neural sentence pair models}}} \\
\hline
Infersent   & 92.8 & 69.7 & 79.6 & 87.8 &  74.0 & 80.3 \\ 

ESIM  & 91.5  & 71.2 &  80.0  & 82.5 & 73.7 & 77.8 \\ 

BERTScore  & 90.6 & 76.5 &  83.0  & 83.2 & 74.3  & 78.5 \\ 

BERT$_{embedding}$  & 84.7 & 53.0 &  65.2  & 77.0 & 74.7  & 75.8 \\ 

BERT$_{finetune}$   & 93.3 & 84.3 & 88.6 &  90.2 & 80.0 & 84.8  \\ 

$\hspace{2.1em}+$ ParaAlign  & 98.4 & 84.2 &  90.7 & 91.9 & 79.0 & 85.0  \\

\hline
\multicolumn{7}{l}{\textbf{\textit{Neural CRF aligner}}} \\
\hline

Our CRF Aligner    & 96.5 & 90.1 &  93.2  & 88.6 & 87.7 & 88.1 \\ 
$\hspace{0.1em}+$ gold ParaAlign   & 97.3 & 91.1 &  94.1 & 88.9   & 88.0 & 88.4  \\ 

\hline

\end{tabular}
\vspace{-0.1cm}
\setlength{\belowcaptionskip}{-14pt}
\caption{Ablation study of our aligner on dev set.}
\label{table:dev_perfprmance}
\end{table}

In this section, we present experiments that compare our neural sentence alignment against the state-of-the-art approaches on {\sc Newsela-Manual} (\S \ref{subsec:newsela_data})  and {\sc Wiki-Manual} (\S \ref{subsec:wiki_data_collection}) datasets. 

\subsection{Existing Methods}
\label{sec:exisitng_aligners}
We compare our neural CRF aligner with the following baselines and state-of-the-art approaches:

\begin{enumerate}[topsep=2pt,itemsep=0pt,partopsep=2pt,parsep=1pt]

    \item Three similarity-based methods:  \textbf{Jaccard similarity} \cite{Xu-EtAl:2015:TACL}, \textbf{TF-IDF} cosine similarity \cite{paetzold-etal-2017-massalign} and a 
    \textbf{logistic regression classifier} trained on our data with lexical features from \citeauthor{STAJNER18.630} \shortcite{STAJNER18.630}.

    \item \textbf{JaccardAlign} \cite{Xu-EtAl:2015:TACL}, which uses Jaccard coefficient for sentence similarity and a greedy approach for alignment.

    \item \textbf{MASSAlign} \cite{paetzold-etal-2017-massalign}, which combines TF-IDF cosine similarity with a vicinity-driven dynamic programming algorithm for alignment.

    \item \textbf{CATS} toolkit \cite{STAJNER18.630}, which uses character n-gram features for sentence similarity and a greedy alignment algorithm. 
\end{enumerate}

\subsection{Evaluation Metrics}

 We report \textbf{Precision}, \textbf{Recall} and \textbf{F1} on two binary classification tasks: \textit{aligned} + \textit{partially-aligned} vs.  \textit{not-aligned} (\textbf{Task 1}) and \textit{aligned} vs. \textit{partially-aligned} + \textit{not-aligned} (\textbf{Task 2}). It should be noted that we excluded identical sentence pairs in the evaluation as they are trivial to classify.  

\subsection{Results}
\label{sec:exp_sent_align_wiki}

Table \ref{table:result_on_testset} shows the results on {\sc Newsela-Manual} test set.  For similarity-based methods, we choose a threshold based on the maximum F1 on the dev set. Our neural CRF aligner outperforms the state-of-the-art approaches by more than 5 points in F1. In particular, our method performs better than the previous work on partial alignments, which contain many interesting simplification operations, such as sentence splitting and paraphrasing with deletion.

Similarly, our CRF alignment model  achieves 85.1 F1 for Task 1 (\textit{aligned} + \textit{partially-aligned} vs.  \textit{not-aligned}) on the {\sc Wiki-Manual} test set. It outperforms one of the previous SOTA approaches CATS \cite{STAJNER18.630} by 15.1 points in F1. We provide more details in Appendix \ref{appendix:alignment_on_wiki}.

\subsection{Ablation Study}
We analyze the design choices crucial for the good performance of our alignment model, namely CRF component, the paragraph alignment and the BERT-based semantic similarity measure. Table \ref{table:dev_perfprmance} shows the importance of each component with a series of ablation experiments on the dev set.

\paragraph{CRF Model}

Our aligner achieves 93.2 F1 and 88.1 F1 on Task 1 and 2, respectively, which is around 3 points higher than its variant without the CRF component (BERT$_{finetune}$ $+$ ParaAlign). Modeling alignment label transitions and sequential predictions helps our neural CRF aligner to handle sentence splitting cases better, especially when sentences undergo dramatic rewriting.

\paragraph{Paragraph Alignment}
\label{sec:analysis_on_paragraph_alignment}
Adding paragraph alignment (BERT$_{finetune}$ $+$ ParaAlign) improves the precision on Task 1 from 93.3 to 98.4 with a negligible decrease in recall when compared to not aligning paragraphs (BERT$_{finetune}$).  Moreover, paragraph alignments generated by our algorithm (Our Aligner) perform close to the gold alignments (Our Aligner $+$ gold ParaAlign) with only 0.9 and 0.3 difference in F1 on Task 1 and 2, respectively.

\paragraph{Semantic Similarity} BERT$_{finetune}$ performs better than other neural models, 
including Infersent \cite{conneau-etal-2017-supervised}, ESIM \cite{chen2016enhanced}, BERTScore \cite{bert-score} and pre-trained BERT embedding \cite{devlin2018bert}. For BERTScore, we use idf weighting, and treat simple sentence as reference.

\section{Experiments on Automatic Sentence Simplification}
\label{sec:experiment_on_ts}

\begin{table}[t!]

\newcolumntype{C}[1]{>{\centering\arraybackslash}p{#1}}
\small
\centering
\begin{tabular}{p{2.99cm}|C{0.6cm}C{0.70cm}|C{0.6cm}C{0.7cm}}
\hline
& \multicolumn{2}{c|}{\textbf{Newsela}} & \multicolumn{2}{c}{\textbf{Wikipedia}} \\
& \textbf{Auto} & \textbf{Old} & \textbf{Auto} & \textbf{Old} \\
\hline

\# of article pairs &  13k & 7.9k  & 138k & 65k \\
\# of sent. pairs (train)  & 394k & 94k & 488k & 298k  \\
\# of sent. pairs (dev)  & 43k & 1.1k & 2k & 2k \\
\# of sent. pairs (test)  & 44k & 1k & 359 & 359 \\
avg. sent. len (complex)  & 25.4  & 25.8  & 26.6 & 25.2\\
avg. sent. len (simple)   & 13.8 & 15.7  & 18.7 & 18.5\\
\hline
\end{tabular}
\small
\setlength{\belowcaptionskip}{-14pt}
\caption{Statistics of our newly constructed parallel corpora for sentence simplification compared to the old datasets \cite{Xu-EtAl:2015:TACL,zhang-lapata-2017-sentence}.}
\label{table:simp_data_stats}
\end{table}

\begin{table*}[ht!]
\small
\centering
\begin{tabular}{l|cccccc|cccccc}
\hline
\multirow{2}{*}{} 
& \multicolumn{6}{c|}{\textbf{Evaluation on our new test set}} & \multicolumn{6}{c}{\textbf{Evaluation on old test set}}  \\

& \textbf{SARI} & \textbf{add} & \textbf{keep}  & \textbf{del} & \textbf{FK} & \textbf{Len} & \textbf{SARI} & \textbf{add} & \textbf{keep}  & \textbf{del} & \textbf{FK} & \textbf{Len} \\ \hline 

Complex (input) & 11.9 & 0.0  & 35.5 & 0.0 & 12 & 24.3 & 12.5 &  0.0 & 37.7 & 0.0 & 11 & 22.9 \\

\hline
\multicolumn{13}{l}{\textbf{\textit{Models trained on old dataset}} (original {\sc Newsela} corpus released in \cite{Xu-EtAl:2015:TACL})} \\
\hline

Transformer$_{rand}$ & 33.1 & 1.8 & 22.1 & 75.4 & 6.8 & 14.2 & 34.1 &  2.0 & 25.5 & \textbf{74.8} & 6.7 & 14.2 \\ 

LSTM  & 35.6 & 2.8 & \textbf{32.1} & 72.0 & 8.2 & 16.9 & 36.2 &  2.5 & \textbf{34.9} & 71.3 & 7.7 & 16.3 \\ 

EditNTS & 35.5 & 1.8 & 30.0 & 75.4 & 7.1 & \underline{14.1} & 36.1 & 1.7 & 32.8 & 73.8 & 7.0 & 14.1  \\ 

Transformer$_{bert}$ & 34.4 & 2.4  & 25.2 & \textbf{75.8} & 7.0 & 14.5 & 35.1 &  2.7 & 27.8 & \textbf{74.8} & 6.8 & 14.3   \\ 

\hline
\multicolumn{13}{l}{\textit{\textbf{Models trained on our new dataset}} ({\sc Newsela-Auto})} \\
\hline

Transformer$_{rand}$ & 35.6 & 3.2 & 28.4 & 75.0 & 7.1 & 14.4 & 35.2 &  2.5 & 29.7 & 73.5 & 7.0 & 14.2 \\ 

LSTM  & \underline{35.8} & \underline{3.9} & 30.5 & 73.1 & 7.0 & 14.3 & \underline{36.4} &  \underline{3.3} & 33.0 & 72.9 & \underline{6.6} & 14.0 \\ 

EditNTS & \underline{35.8} & 2.4 & 29.4 & \underline{75.6} & \underline{6.3} & 11.6  & 35.7 & 1.8 & 31.1 & \underline{74.2} & \textbf{6.1} & \underline{11.5}  \\ 

Transformer$_{bert}$ & \textbf{36.6} &\textbf{4.5}  & \underline{31.0} & 74.3 & \textbf{6.8} & \textbf{13.3} & \textbf{36.8} &  \textbf{3.8} & \underline{33.1} & 73.4 & 6.8 & \textbf{13.5}  \\ 

\hline
Simple (reference) & -- & -- & -- & -- & 6.6 & 13.2 & -- & -- & -- & -- & 6.2 & 12.6 \\
\hline
\end{tabular}
\setlength{\belowcaptionskip}{-10pt}
\caption{Automatic evaluation results on {\sc Newsela} test sets comparing models trained on our dataset {\sc Newsela-Auto} against the existing dataset \cite{Xu-EtAl:2015:TACL}. We report \textbf{SARI, the main automatic metric} for simplification, precision for deletion and F1 scores for adding and keeping operations. Add scores are low partially because we are using one reference. \textbf{Bold} typeface and \underline{underline} denote the best and the second best performances respectively. For Flesch-Kincaid (FK) grade level and average sentence length (Len), we consider the values closest to reference as the best.}
\label{table:newsela}
\end{table*}

\begin{table}[t!]
\small
\centering
\begin{tabular}{l|cccc}
\hline
\textbf{Model} & \textbf{F} & \textbf{A} & \textbf{S} & \textbf{Avg.} \\ \hline 

LSTM & 3.44 & 2.86 & 3.31 & 3.20   \\   

EditNTS \cite{dong-etal-2019-editnts}$^\dagger$ & 3.32 & 2.79 & \textbf{3.48} & 3.20  \\   

Rerank \cite{kriz-etal-2019-complexity}$^\dagger$ & 3.50 & 2.80 & 3.46 & 3.25 \\ 

Transformer$_{bert}$ (this work) & \textbf{3.64} & \textbf{3.12} & 3.45 & \textbf{3.40} \\ 
\hline
Simple (reference) & 3.98 & 3.23 & 3.70 & 3.64  \\
\hline
\end{tabular}
\setlength{\belowcaptionskip}{-2pt}
\caption{Human evaluation of fluency (\textbf{F}), adequacy (\textbf{A}) and simplicity (\textbf{S}) on the old {\sc Newsela} test set. $\dagger$We used the system outputs shared by the authors.}
\label{table:human_eval_sota}
\end{table}

In this section, we compare different automatic text simplification models trained on our new parallel corpora, {\sc Newsela-Auto}  and {\sc Wiki-Auto}, with their counterparts trained on the existing datasets. We establish a new state-of-the-art for sentence simplification by training a Transformer model with initialization from pre-trained BERT checkpoints.  



\subsection{Comparison with existing datasets}

Existing datasets of complex-simple sentences,  {\sc Newsela} \cite{Xu-EtAl:2015:TACL} and {\sc WikiLarge} \cite{zhang-lapata-2017-sentence}, were aligned using lexical similarity metrics. {\sc Newsela} dataset \cite{Xu-EtAl:2015:TACL} was aligned using JaccardAlign (\S \ref{sec:exisitng_aligners}).
{\sc WikiLarge} is a concatenation of three early datasets \cite{zhu-etal-2010-monolingual,woodsend-lapata-2011-learning, coster-kauchak-2011-simple} where sentences in Simple/Normal English Wikipedia and editing history were aligned by TF-IDF cosine similarity. 



For our new {\sc Newsela-Auto}, we partitioned the article sets such that there is no overlap between the new train set and the old test set, and vice-versa. Following \citeauthor{zhang-lapata-2017-sentence} \shortcite{zhang-lapata-2017-sentence}, we also excluded sentence pairs corresponding to the levels 0--1, 1--2 and 2--3. Similar to \cite{stajner-etal-2015-deeper}, for our {\sc Wiki-Auto} dataset, we eliminated sentence pairs with high ($>$0.9) or low ($<$0.1) lexical overlap based on GLEU scores \cite{Wu2016GooglesNM}. We observed that sentence pairs with low GLEU are often inaccurate paraphrases with only shared named entities and the pairs with high GLEU are dominated by sentences merely copied without simplification. We used the benchmark {\sc Turk} corpus \cite{xu-etal-2016-optimizing} for evaluation on Wikipedia, which consists of 8 human-written references for sentences in the validation and test sets. We discarded sentences in \textsc{Turk} corpus from \textsc{Wiki-auto}. Table \ref{table:simp_data_stats} shows the statistics of the existing and our new datasets. 







\subsection{Baselines and Simplification Models}
We compare the following seq2seq models trained using our new datasets versus the existing datasets: 

\begin{enumerate}[topsep=2pt,itemsep=0pt,partopsep=2pt,parsep=1pt]

\item A \textbf{BERT-initialized Transformer}, where the encoder and decoder follow the  BERT$_{base}$ architecture. The encoder is initialized with the same checkpoint and the decoder is randomly initialized \cite{sascha2019}. 

\item A \textbf{randomly initialized Transformer} with the same BERT$_{base}$ architecture as above.

\item A \textbf{BiLSTM-based encoder-decoder} model used in \citeauthor{zhang-lapata-2017-sentence} \shortcite{zhang-lapata-2017-sentence}.

\item \textbf{EditNTS} \cite{dong-etal-2019-editnts},\footnote{https://github.com/yuedongP/EditNTS} a state-of-the-art neural programmer-interpreter \cite{DBLP:journals/corr/ReedF15} approach that predicts explicit edit operations sequentially.

\end{enumerate}

In addition, we compared our BERT-initialized Transformer model with the released system outputs from \citeauthor{kriz-etal-2019-complexity} \shortcite{kriz-etal-2019-complexity} and EditNTS \cite{dong-etal-2019-editnts}. We implemented our LSTM and Transformer models using Fairseq.\footnote{https://github.com/pytorch/fairseq} We provide the model and training details in Appendix \ref{appendix:imp}. 

\subsection{Results}
In this section, we evaluate different simplification models trained on our new datasets versus on the old existing datasets using both automatic and human evaluation.

\begin{table}[t!]
\small
\centering
\begin{tabular}{l|c|cccc}
\hline
\textbf{Model} & \textbf{Train} & \textbf{F} & \textbf{A} & \textbf{S} & \textbf{Avg.} \\ \hline 

LSTM & old & 3.57 & \textbf{3.27} & 3.11 & 3.31   \\   

LSTM & new & 3.55 & 2.98 & 3.12 & 3.22  \\

Transformer$_{bert}$ & old & 2.91 & 2.56 & 2.67 & 2.70 \\ 

Transformer$_{bert}$ & new & \textbf{3.76}  & 3.21 & \textbf{3.18} & \textbf{3.39} \\ 
\hline
Simple (reference) & --- & 4.34 & 3.34  &  3.37 & 3.69  \\
\hline
\end{tabular}
\setlength{\belowcaptionskip}{-12pt}
\caption{Human evaluation of fluency (\textbf{F}), adequacy (\textbf{A}) and simplicity (\textbf{S}) on {\sc Newsela-Auto} test set.}
\label{table:human_eval_old_new}
\end{table}

\subsubsection{Automatic Evaluation}

We report \textbf{SARI} \cite{xu-etal-2016-optimizing}, Flesch-Kincaid (\textbf{FK}) grade level readability \cite{kincaid}, and average sentence length (\textbf{Len}). While SARI compares the generated sentence to a set of reference sentences in terms of correctly inserted, kept and deleted n-grams $(n \in \{1, 2, 3, 4\})$, FK measures the readability of the generated sentence. We also report the three rewrite operation scores used in SARI: the precision of delete (\textbf{del}), the F1-scores of add (\textbf{add}), and keep (\textbf{keep})  operations. 

\begin{figure}[t!]
\centering
\includegraphics[width=0.4\textwidth]{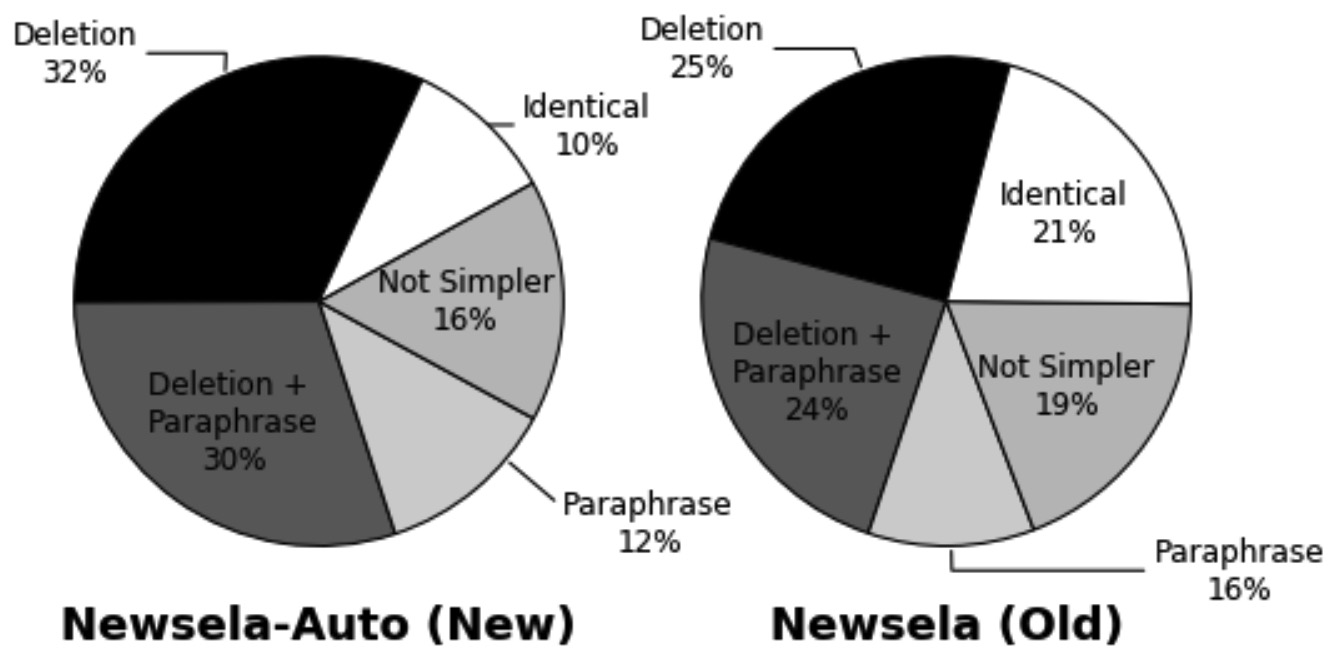}
\setlength{\belowcaptionskip}{-14pt}
\caption{Manual inspection of 100 random sentences generated by Transformer$_{bert}$ trained on {\sc Newsela-Auto} and existing {\sc Newsela} datasets, respectively.}
\label{fig:simplificaiton_operation_stat}
\end{figure}

Tables \ref{table:newsela} and \ref{table:wiki} show the results on Newsela and Wikipedia datasets respectively. Systems trained on our datasets outperform their equivalents trained on the existing datasets according to SARI. The difference is notable for Transformer$_{bert}$ with a 6.4\% and 3.7\% increase in SARI on {\sc Newsela-Auto} test set and {\sc Turk} corpus, respectively. Larger size and improved quality of our datasets enable the training of complex Transformer models. In fact, Transformer$_{bert}$ trained on our new datasets outperforms the existing state-of-the-art systems for automatic text simplification. Although improvement in SARI is modest for LSTM-based models (LSTM and EditNTS), the increase in F1 scores for addition and deletion operations indicate that the models trained on our datasets make more meaningful changes to the input sentence.

\subsubsection{Human Evaluation}

We also performed human evaluation by asking five Amazon Mechanical Turk workers to rate  fluency, adequacy and simplicity (detailed instructions in Appendix \ref{appendix:human_eval}) of 100 random sentences generated by different simplification models trained on {\sc Newsela-Auto} and the existing dataset. Each worker evaluated these aspects on a 5-point Likert scale. We averaged the ratings from five workers. Table \ref{table:human_eval_old_new} demonstrates that Transformer$_{bert}$ trained on {\sc Newsela-Auto} greatly outperforms the one trained on the old dataset. Even with shorter sentence outputs, our Transformer$_{bert}$ retained similar adequacy as the LSTM-based models. Our Transformer$_{bert}$ model also achieves better fluency, adequacy, and overall ratings compared to the SOTA systems (Table \ref{table:human_eval_sota}). We provide examples of system outputs in Appendix \ref{appendix:system_outputs}. Our manual inspection (Figure \ref{fig:simplificaiton_operation_stat}) also shows that Transfomer$_{bert}$ trained on {\sc Newsela-Auto} performs 25\% more paraphrasing and deletions than its variant trained on the previous {\sc Newsela} \cite{Xu-EtAl:2015:TACL} dataset.

\begin{table}[t!]
\small
\centering
\begin{tabular}{l|cccccc}
\hline
& \textbf{SARI} & \textbf{add} & \textbf{keep}  & \textbf{del} & \textbf{FK} & \textbf{Len} \\ \hline 
Complex (input) & 25.9 & 0.0 & 77.8 & 0.0 & 13.6 & 22.4 \\
\hline
\multicolumn{7}{l}{\textbf{\textit{Models trained on old dataset}} ({\sc WikiLarge})} \\
\hline
LSTM  & 33.8 & 2.5 & 65.6 & 33.4 & \underline{11.6} & \underline{20.6}  \\
Transformer$_{rand}$ & 33.5 & 3.2 & 64.1 & 33.2 & 11.1 & 17.7 \\ 
EditNTS & 35.3 & 3.0 & 63.9 & \underline{38.9} & 11.1 & 18.5  \\ 
Transformer$_{bert}$ & 35.3 & \underline{4.4} & 66.0 & 35.6 & 10.9 & 17.9  \\ 
\hline
\multicolumn{7}{l}{\textit{\textbf{Models trained on our new dataset}} ({\sc Wiki-Auto})} \\
\hline
LSTM  & 34.0 & 2.8 & 64.0 & 35.2 & 11.0 & 19.3\\ 
Transformer$_{rand}$ & 34.7 & 3.3 & \textbf{68.8} & 31.9 & \textbf{11.7} & 18.7 \\ 
EditNTS & \underline{36.4} & 3.6 & 66.1 & \textbf{39.5} & \underline{11.6} & \textbf{20.2}   \\
Transformer$_{bert}$ & \textbf{36.6} & 
\textbf{5.0} & \underline{67.6} & 37.2 & 11.4 & 18.7 \\ 
\hline
Simple (reference) &  -- & -- & -- & -- & 11.7 & 20.2 \\
\hline
\end{tabular} 
\setlength{\belowcaptionskip}{-8pt}
\caption{Automatic evaluation results on Wikipedia {\sc Turk} corpus comparing models trained on {\sc Wiki-Auto} and {\sc WikiLarge} \cite{zhang-lapata-2017-sentence}.}
\label{table:wiki}
\end{table}

\section{Related Work}
 \textbf{Text simplification} is considered as a  text-to-text generation task where the system learns how to simplify from complex-simple sentence pairs. There is a long line of research using methods based on hand-crafted rules \cite{siddharthan2006syntactic,niklaus-etal-2019-transforming}, statistical machine translation \cite{narayan-gardent-2014-hybrid,xu-etal-2016-optimizing, wubben-etal-2012-sentence}, or neural seq2seq models \cite{zhang-lapata-2017-sentence, zhao-etal-2018-integrating, nisioi-etal-2017-exploring}.  As the existing datasets were built using lexical similarity metrics, they frequently omit paraphrases and sentence splits. While training on such datasets creates conservative systems that rarely paraphrase, evaluation on these datasets exhibits an unfair preference for deletion-based simplification over paraphrasing. 
 
 \vspace{.1cm}
 
 \noindent\textbf{Sentence alignment} has been widely used to extract complex-simple sentence pairs from parallel articles for training text simplification systems.  Previous work used surface-level similarity metrics, such as TF-IDF cosine similarity \cite{zhu-etal-2010-monolingual,woodsend-lapata-2011-learning, coster-kauchak-2011-simple, paetzold-etal-2017-massalign}, Jaccard-similarity  \cite{Xu-EtAl:2015:TACL}, and other lexical features \cite{hwang-EtAl:2015:NAACL-HLT,STAJNER18.630}. Then, a greedy \cite{STAJNER18.630} or dynamic programming \cite{barzilay-elhadad-2003-sentence,paetzold-etal-2017-massalign} algorithm was used to search for the optimal alignment. Another related line of research  \cite{smith-etal-2010-extracting, tufis-etal-2013-wikipedia, tsai-roth-2016-cross, gottschalk-acm-2017-multiwiki, aghaebrahimian2018deep,thompson-koehn-2019-vecalign} aligns parallel sentences in bilingual corpora for machine translation.

\section{Conclusion}

In this paper, we proposed a novel neural CRF model for sentence alignment, which substantially outperformed the existing approaches. We created two high-quality manually annotated datasets ({\sc Newsela-Manual} and {\sc Wiki-Manual}) for training and evaluation. Using the neural CRF sentence aligner, we constructed two largest sentence-aligned datasets to date ({\sc Newsela-Auto} and {\sc Wiki-Auto}) for text simplification. We showed that a BERT-initalized Transformer trained on our new datasets establishes new state-of-the-art performance for automatic sentence simplification.

\section*{Acknowledgments}

We thank three anonymous
reviewers for their helpful comments, Newsela for sharing the data, Ohio Supercomputer Center \cite{Oakley2012} and NVIDIA for providing GPU computing resources. We also thank Sarah Flanagan, Bohan Zhang, Raleigh Potluri, and Alex Wing for help with data annotation. This research is supported in part by the NSF awards IIS-1755898 and IIS-1822754, ODNI and IARPA via the BETTER program contract 19051600004, ARO and DARPA via the SocialSim program contract W911NF-17-C-0095, Figure Eight AI for Everyone Award, and Criteo Faculty Research Award to Wei Xu. The views and conclusions contained herein are those of the authors and should not be interpreted as necessarily representing the official policies, either expressed or implied, of NSF, ODNI, IARPA, ARO, DARPA or the U.S. Government. The U.S. Government is authorized to reproduce and distribute reprints for governmental purposes notwithstanding any copyright annotation therein.

\newpage

\bibliography{acl2020}
\bibliographystyle{acl_natbib}
\clearpage

\appendix

\section{Neural CRF Alignment Model}
\subsection{Implementation Details}
\label{appendix:details_of_alignment}
We used  PyTorch\footnote{https://pytorch.org/} to implement our neural CRF alignment model. For the sentence encoder, we used Huggingface implementation\cite{Wolf2019HuggingFacesTS} of BERT$_{base}$ \footnote{https://github.com/google-research/bert} architecture with 12 layers of Transformers. When  fine-tuning the BERT model, we use the representation of   {\ttfamily [CLS]} token for classification. We use cross entropy loss and update the weights in all layers. Table \ref{table:crf_model_details} summarizes the hyperparameters of  our model. Table \ref{table:paragraph_alignment_thresholds_on_newsela} provides the thresholds for our paragraph alignment Algorithm \ref{alg:paragraph_alignment}, which were chosen based on \textsc{Newsela-Manual} dev data.

\begin{table}[ht!]
\small
\centering
\begin{tabular}{rl|r l}
\hline
\textbf{Parameter} & \textbf{Value} &\textbf{Parameter} & \textbf{Value}  \\
\hline
hidden units &  768 & \# of layers & 12\\
learning rate & 0.00002 & \# of heads & 12 \\
 max sequence length & 128 & batch size & 8\\
\hline
\end{tabular}
\small
\caption{Parameters of our neural CRF sentence alignment model.}
\label{table:crf_model_details}
\end{table}

\begin{table}[ht!]
\small
\centering
\begin{tabular}{c| l}
\hline
\textbf{Threshold} & \textbf{Value}  \\
\hline
$\tau_1$ &  0.1 \\
$\tau_2$ & 0.34 \\
$\tau_3$ & 0.9998861788416304  \\
$\tau_4$ & 0.998915818299745 \\
$\tau_5$ & 0.5 \\
\hline
\end{tabular}
\small
\caption{The thresholds in paragraph alignment Algorithm \ref{alg:paragraph_alignment} for Newsela data.}
\label{table:paragraph_alignment_thresholds_on_newsela}
\end{table}

For Wikipedia data, we tailored our paragraph alignment algorithm (Algorithm \ref{alg:paragraph_similarity_wiki} and \ref{alg:paragraph_alignment_wiki}). Table \ref{table:paragraph_alignment_thresholds_on_wiki} provides the thresholds for Algorithm \ref{alg:paragraph_alignment_wiki}, which were chosen based on \textsc{Wiki-Manual} dev data.

\begin{table}[h!]
\small
\centering
\begin{tabular}{c| l}
\hline
\textbf{Threshold} & \textbf{Value}  \\
\hline
$\tau_1$ &  0.991775706637882 \\
$\tau_2$ & 0.8 \\
$\tau_3$ & 0.5  \\
$\tau_4$ & 5 \\
$\tau_5$ & 0.9958 \\
\hline
\end{tabular}
\small
\caption{The thresholds in paragraph alignment Algorithm \ref{alg:paragraph_alignment_wiki} for Wikipedia data.}
\label{table:paragraph_alignment_thresholds_on_wiki}
\end{table}

\begin{algorithm}[ht!]
\small
\SetKwInput{kwInit}{Initialize}

\SetAlgoLined
\kwInit{$simP$ $\in$ $\mathbb{R}^{1 \times k \times l}$ to $0^{1 \times k \times l}$}
\For{$i\leftarrow 1$ \KwTo $k$}
{
    \For{$j\leftarrow 1$ \KwTo $l$}
    {

        \resizebox{1.0\hsize}{!}{
            $ simP[1, i, j] = \max\limits_{s_p \in S_i, c_q \in C_j} simSent(s_p, c_q)$ \nonumber
        }
        
    }

}
\Return  $simP$
\caption{Pairwise Paragraph Similarity}
\label{alg:paragraph_similarity_wiki}
\end{algorithm}

\begin{algorithm}[ht]
\small
\SetKwInOut{Input}{Input}
\SetKwInput{kwInit}{Initialize}

\SetAlgoLined
\Input{$simP$ $\in$ $\mathbb{R}^{1\times k \times l}$}
\kwInit{$alignP$ $\in$ $\mathbb{I}^{k \times l}$ to $0^{k \times l}$}
\For{$i\leftarrow 1$ \KwTo $k$}
{

    $cand$ = [] \\ 
    
    \For{$j\leftarrow 1$ \KwTo $l$}{

        \If{$simP[1, i,j] > \tau_1$  \& $d(i, j) < \tau_2$ } {$cand.append(j)$}

    }
    $range = max(cand) - min(cand)$ \\ 
        \If{len$(cand) > 1$ \& $ range / l  > \tau_3 $ \& $range > \tau_4$}{
        
        $dist = []$ \\
        \For{$m \in cand $ }{$dist.append(abs(m - i))$}

        $ j_{cloest} = cand[\argmin\limits_{n}  dist[n]] $ \\

        \For{$m \in cand $ }{\If{$m\neq j_{cloest} \& simP[1, i,m] \leq \tau_5 $}{$cand.remove(m)$}}}
    \For{$m \in cand $ }{$alignP[i, m]=1$}
}
\Return  $alignP$
\caption{Paragraph Alignment Algorithm}
\label{alg:paragraph_alignment_wiki}
\end{algorithm}

\section{Sentence Aligned Wikipedia Corpus}
\label{appendix:details_of_processing_wikipedia}
We present more details about our pre-processing steps for creating the {\sc Wiki-Manual} and {\sc Wiki-Auto} corpora here. In Wikipedia, Simple English is considered as a language by itself. When extracting articles from  Wikipedia dump, we removed the meta-page and disambiguation pages. We also removed sentences with less than 4 tokens and sentences that end with a colon.

After the pre-processing and matching steps, there are 13,036 article pairs in which the simple article contains only one sentence. In most cases, that one sentence is aligned to the first sentence in the complex article. However, we find that the patterns of these sentence pairs are very repetitive (e.g., XXX is a city in XXX. XXX is a football player in XXX.). Therefore, we use regular expressions to filter out the sentences with repetitive patterns.  Then, we use a BERT model fine-tuned on the {\sc Wiki-Manual} dataset to compute the semantic similarity of each sentence pair and keep the ones with a similarity larger than a threshold tuned on the dev set. After filtering, we ended up with 970 aligned sentence pairs in total from these 13,036 article pairs.

\section{Sentence Alignment on Wikipedia}
\label{appendix:alignment_on_wiki}

In this section, we compare different  approaches for sentence alignment on the \textsc{Wiki-Manual} dataset. Tables \ref{table:dev_perfprmance_wikipedia} and \ref{table:test_perfprmance_wikipedia} report the performance for Task 1 (\textit{aligned} + \textit{partially-aligned} vs. \textit{not-aligned}) on dev and test set. To generate prediction for MASSAlign, CATS and two BERT$_{finetune}$ methods, we first utilize the method in \S \ref{subsec:wiki_data_collection} to select candidate sentence pairs, as we found this step helps to improve their accuracy. Then we apply the similarity metric from each model to calculate the similarity of each candidate sentence pair. We tune a threshold for max f1 on the dev set and apply it to the test set. Candidate sentence pairs with a similarity larger than the threshold will be predicted as \textit{aligned}, otherwise \textit{not-aligned}. Sentence pairs that are not selected as candidates will also be predicted as \textit{not-aligned}.

\begin{table}[pht]
\small
\centering
\begin{tabular}{l|ccc}
\hline
\multirow{2}{*}{} & \multicolumn{3}{c}{\textbf{Dev set}} \\
 & \multicolumn{1}{c}{\textbf{P}}  & \multicolumn{1}{c}{\textbf{R}} &\multicolumn{1}{c}{\textbf{F}}   
 \\ 

\hline

MASSAlign \cite{paetzold-etal-2017-massalign}   & 72.9 & 79.5 & 76.1  \\ 

CATS \cite{STAJNER18.630}  & 65.6 & 82.7 & 73.2  \\ 
\hline
BERT$_{finetune}$ ({\sc Newsela-Manual})  &  82.6 & 83.9 & 83.2 \\
BERT$_{finetune}$ ({\sc Wiki-Manual})  & 87.9 & 85.4 & 86.6 \\ 
$\hspace{2.1em}+$ ParaAlign   & 88.6 & 85.4 & 87.0 \\ 
Our CRF Aligner ({\sc Wiki-Manual})  &    92.4  & 85.8 & 89.0 \\

\hline
\end{tabular}

\caption{Performance  of different sentence alignment methods on the {\sc Wiki-Manual} dev set for Task 1.}

\label{table:dev_perfprmance_wikipedia}
\end{table}
\begin{table}[pht]
\small
\centering
\begin{tabular}{l|ccc}
\hline
\multirow{2}{*}{} & \multicolumn{3}{c}{\textbf{Test set}} \\
 & \multicolumn{1}{c}{\textbf{P}}  & \multicolumn{1}{c}{\textbf{R}} &\multicolumn{1}{c}{\textbf{F}}   
 \\ 

\hline

MASSAlign \cite{paetzold-etal-2017-massalign}   & 68.6 & 72.5 & 70.5  \\ 

CATS \cite{STAJNER18.630}  & 68.4 & 74.4 & 71.3  \\
\hline
BERT$_{finetune}$ ({\sc Newsela-Manual}) & 80.6 & 78.8 & 79.6 \\
BERT$_{finetune}$ ({\sc Wiki-Manual}) & 86.3 & 82.4 & 84.3 \\
$\hspace{2.1em}+$ ParaAlign   & 86.6 & 82.4 & 84.5 \\ 
Our CRF Aligner ({\sc Wiki-Manual})  &    89.3  & 81.6 & 85.3 \\ 

\hline
\end{tabular}

\caption{Performance  of different sentence alignment methods on the {\sc Wiki-Manual} test set for Task 1.}

\label{table:test_perfprmance_wikipedia}
\end{table}

\section{Sentence Simplification}
\subsection{Implementation Details}
\label{appendix:imp}
We used Fairseq\footnote{https://github.com/pytorch/fairseq} toolkit to implement our Transformer \cite{VaswaniSPUJGKP17} and LSTM \cite{hochreiter1997long} baselines.  For the Transformer baseline, we followed BERT$_{base}$ \footnote{https://github.com/google-research/bert} architecture for both encoder and decoder. We initialized the encoder using BERT$_{base}$ uncased checkpoint. \citeauthor{sascha2019} \shortcite{sascha2019}   used a similar model for sentence fusion and summarization. We trained each model using Adam optimizer with a learning rate of 0.0001, linear learning rate warmup of 40k steps and 200k training steps.  We tokenized the data with BERT WordPiece tokenizer. Table \ref{table:transformer_parameters}  shows the values of other hyperparameters. 

For the LSTM baseline, we replicated the LSTM  encoder-decoder model used by  \citeauthor{zhang-lapata-2017-sentence} \shortcite{zhang-lapata-2017-sentence}. We  preprocessed the data by replacing the named entities in a sentence using spaCy\footnote{https://spacy.io/} toolkit. We also replaced all the words with frequency less than three with {\ttfamily <UNK>}. If our model predicted {\ttfamily <UNK>}, we replaced it with the aligned source word \cite{jean-etal-2015-montreal}. Table \ref{table:lstm_parameters}  summarizes the hyperparameters of LSTM model. We used 300-dimensional GloVe word embeddings \cite{pennington2014glove}  to initialize the embedding layer.

\begin{table}[t!]
\small
\centering
\begin{tabular}{rl|rl}
\hline
\textbf{Parameter} & \textbf{Value} & \textbf{Parameter} & \textbf{Value} \\
\hline
hidden units &  768 &  batch size & 32 \\
filter size & 3072 &  max len & 100 \\
\# of layers & 12 &  activation & GELU \\
 attention heads & 12 &  dropout & 0.1 \\
 loss & CE & seed &  13 \\
\hline
\end{tabular}
\small
\caption{Parameters of our Transformer model.}
\label{table:transformer_parameters}
\end{table}

\begin{table}[t!]
\small
\centering
\begin{tabular}{rl|rl}
\hline
\textbf{Parameter} & \textbf{Value} & \textbf{Parameter} & \textbf{Value} \\
\hline
hidden units & 256 &  batch size & 64 \\
embedding dim & 300 &  max len & 100 \\
\# of layers & 2 &  dropout & 0.2 \\
 lr & 0.001 &  optimizer &  Adam \\
  clipping & 5 & epochs &  30 \\
 min vocab freq & 3 & seed &  13 \\
\hline
\end{tabular}
\small
\caption{Parameters of our LSTM model.}
\label{table:lstm_parameters}
\end{table}

\clearpage
\subsection{Human Evaluation}
\label{appendix:human_eval}
\noindent\begin{minipage}{\textwidth}
   \small
    \centering
    \includegraphics[width=0.95\textwidth]{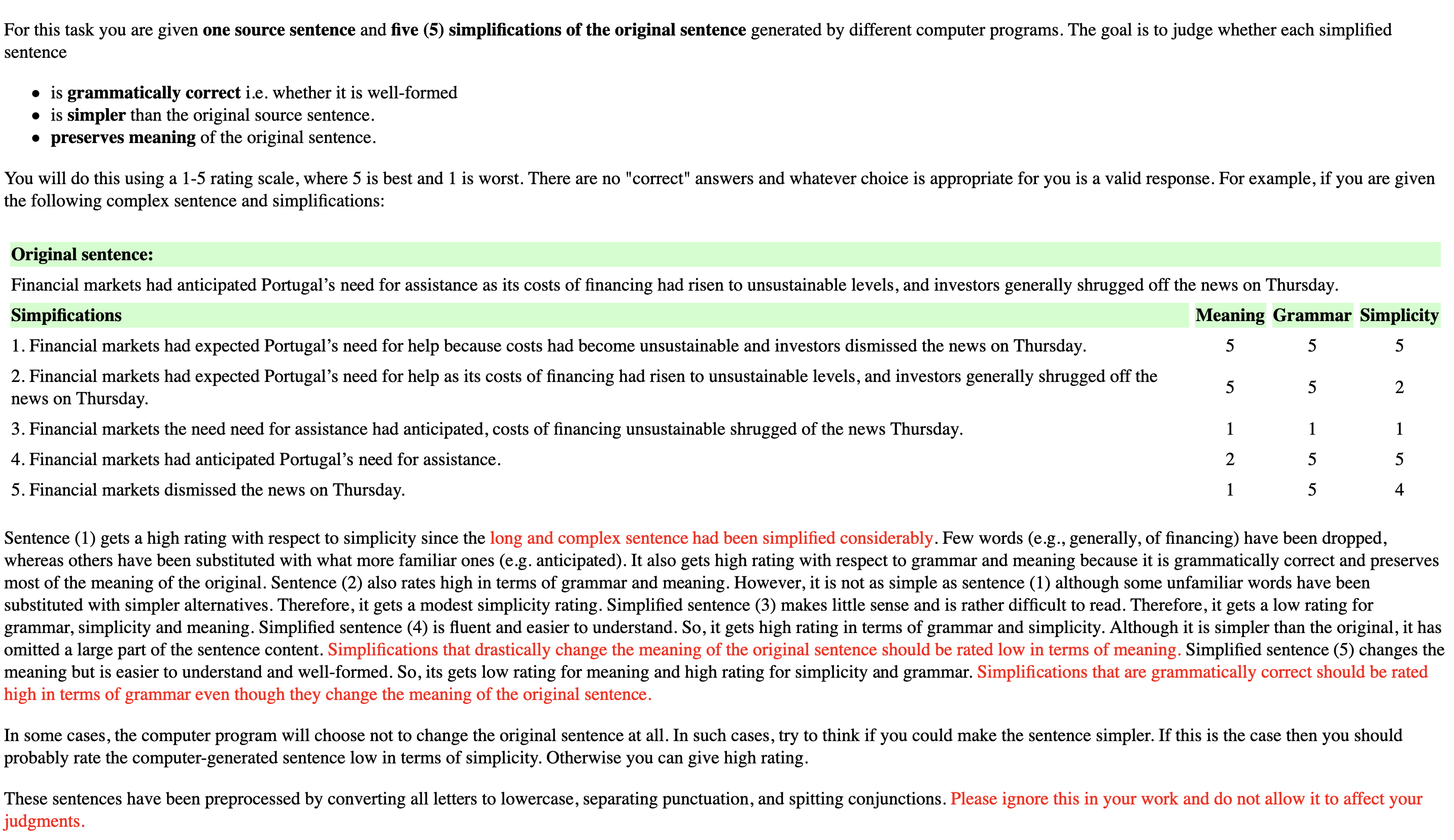}
    \captionof{figure}{Instructions provided to Amazon Mechanical Turk workers to evaluate generated simplified sentences.  We used the same instructions as described in \citeauthor{kriz-etal-2019-complexity} \shortcite{kriz-etal-2019-complexity}.}
    \label{fig:evaluation_interface}
\end{minipage}

\clearpage

\onecolumn
\subsection{Example System Outputs}
\label{appendix:system_outputs}
\vspace{-10pt}

\begin{table}[ht]
\small
\centering
\begin{tabular}{l|p{12.2cm}}
\hline
\multicolumn{2}{c}{\textbf{Examples}} \\ 
\hline 
\multicolumn{2}{l}{\textbf{\textit{Generated by LSTM baseline}}}\\
\hline
Complex (input) & \textit{In Seattle , eight activists between ages 10 and 15 \textbf{petitioned} Washington state last year to \textbf{adopt stricter science-based regulations to protect them against climate change.}} \\ 
Simple (reference) & \textit{In Seattle, eight youths between 10 to 15 years old petitioned the state of Washington to change the law.}\\
New (this work) & \textit{in seattle , eight activists between ages 10 and 15 \textbf{asked} washington state last year \textbf{to keep the environment safe.}} (Phrasal Praphrase + Deletion) \\
Old \cite{Xu-EtAl:2015:TACL} & \textit{in seattle , eight activists between ages 10 and 15 \textbf{asked} washington state last year to adopt stricter science - based rules to protect them against climate change.} (Lexical Paraphrase) \\ 
\hline
Complex (input) & \textit{He \textbf{recognized that} another \textbf{recommendation} would be \textbf{controversial with police groups}: \textbf{independent investigations after police shootings.}} \\ 
Simple (reference) & \textit{He admitted that police would not like one of the recommendations.} \\
New (this work) & \textit{he \textbf{thought another suggestion} would be \textbf{against the police.}} (Phrasal Paraphrase + Deletion)\\
Old \cite{Xu-EtAl:2015:TACL} & \textit{he recognized that another \textbf{suggestion} would be controversial with police groups.} (Lexical Paraphrase +  Deletion)\\ 
\hline

Complex (input) & \textit{The Philadelphia Museum of Art has two \textbf{famous selfie} spots , \textbf{both from the movie " Rocky. "}} \\  
Simple (reference) & \textit{The Philadelphia Museum of Art has two big selfie spots.}\\
New (this work) & \textit{the philadelphia museum of art has two \textbf{picture} spots.} (Lexical Paraphrase + Deletion)\\
Old \cite{Xu-EtAl:2015:TACL} & \textit{the philadelphia museum of art has two famous spots.} (Deletion)\\

\hline 
\multicolumn{2}{l}{\textbf{\textit{Generated by Transformer$_{bert}$}}}\\
\hline

Complex (input) & \textit{Some \textbf{Chicago residents got angry about it.}} \\  
Simple (reference) & \textit{The plan made some people angry.}\\
New (this work) & \textit{some \textbf{people in chicago were angry.}} (Phrasal Paraphrase)\\
Old \cite{Xu-EtAl:2015:TACL} & \textit{some chicago residents got angry.} (Deletion)\\ 
\hline

Complex (input) & \textit{
\textbf{Emissions standards have been tightened} , and the \textbf{government is investing money in solar , wind and other renewable energy}.} \\
Simple (reference) & \textit{China has also put a great deal of money into solar, wind and other renewable energy.} \\
New (this work) & \textit{the \textbf{government is putting aside money for new types of energy.}} (Phrasal Paraphrase + Deletion)\\
Old \cite{Xu-EtAl:2015:TACL} & \textit{the government is investing in \textbf{money} , wind and other \textbf{equipment.}} (Lexical Paraphrase + Deletion)\\ 
\hline

Complex (input) & \textit{On Feb. 9 , 1864 , he was sitting for several portraits , \textbf{including the one used for the \$5 bill.}	}\\
Simple (reference) & \textit{On Feb. 9, 1864, several artists painted pictures of him.}\\
New (this work) & \textit{on feb. 9, 1864 , he was sitting for several portraits.} (Deletion)\\
Old \cite{Xu-EtAl:2015:TACL} & \textit{on feb 9, 1864 , he was sitting for several , including the \$ 5 bill for the bill.} (Deletion)\\ 
\hline
 
\end{tabular}
\setlength{\belowcaptionskip}{-18pt}

\caption{Examples of simplified sentences generated by LSTM and Transformer$_{bert}$ models trained on our new {\sc Newsela-Auto} (this work) and old existing {\sc Newsela} \cite{Xu-EtAl:2015:TACL} datasets. The source sentences are from our new  {\sc Newsela-Auto} test set. Models trained on our new data rephrase the input sentence more often than the models trained on old data. \textbf{Bold} indicates deletions or paraphrases.}
\label{table:simp_operations_egs}
\end{table}

\begin{table}[!htbp]
\small
\centering
\begin{tabular}{l|p{11.3cm}}
\hline
\multicolumn{2}{c}{\textbf{Examples}} \\ 
\hline 
Complex (input) & \textit{Now at age 9, his teachers say Richie reads at the level of a student in high school, and his vocabulary is well above those of his classmates.} \\  
Simple (reference) & \textit{He reads like a high school student.}\\
LSTM & \textit{now \textbf{he is} age 9.} \\ 
EditNTS \cite{dong-etal-2019-editnts} & \textit{he say his classmates \textbf{are using a special job.}} \\ 
Rerank \cite{kriz-etal-2019-complexity} & \textit{\textbf{but it} is well above those of his classmates.} \\ 
Transfomer$_{bert}$ (this work) & \textit{now at age 9 , his teachers say \textbf{that richie reads high schoolwork.}} \\
\hline
Complex (input) & \textit{He can recall the special feeling when, at age 7, he built his first kite and saw it waft into the air.}  \\  
Simple (reference) & \textit{He can remember the special feeling when he built his first kite .}\\
LSTM & \textit{he can \textbf{remember the people} when he was age 7.} \\ 
EditNTS \cite{dong-etal-2019-editnts} &  \textit{, at age 7, he built his first kite.} \\ 
Rerank \cite{kriz-etal-2019-complexity} & \textit{he could \textbf{remember} the special feeling when.}  \\ 
Transfomer$_{bert}$ (this work) & \textit{he can \textbf{remember} the special feeling when he was 7 years old.} \\
\hline
Complex (input)  & \textit{Following the action , two middle-aged brothers slid down a hill holding signs.} \\ 
Simple (reference)  & \textit{For example , two grownup brothers slid down a hill holding signs .} \\ 
LSTM & \textit{$<$unk$>$ middle - aged brothers slid down a hill holding signs.} \\ 
EditNTS \cite{dong-etal-2019-editnts} & \textit{two middle-aged brothers \textbf{, 14 ,} heard down a hill signs.} \\ 
Rerank \cite{kriz-etal-2019-complexity}& \textit{\textbf{he made a} hill holding signs.} \\ 
Transfomer$_{bert}$ (this work) & \textit{two middle-aged brothers slid down a hill holding signs.} \\
\hline
\end{tabular}
\setlength{\belowcaptionskip}{-18pt}

\caption{Examples of simplifications generated by our best model, Transformer$_{bert}$, and other baselines, namely, EditNTS \cite{dong-etal-2019-editnts}, Rerank \cite{kriz-etal-2019-complexity} and LSTM on the old {\sc Newsela} test set. Both LSTM and Transformer$_{bert}$ are trained on {\sc Newsela-Auto}. For EditNTS and Rerank, we use the system outputs shared by their original authors. \textbf{Bold} indicates new phrases introduced by the model.}
\label{table:simp_sys_output_egs}
\end{table}

\clearpage
\section{Annotation Interface}
\subsection{Crowdsourcing Annotation Interface}
\label{appendix:human-annotation}

\noindent\begin{minipage}{\textwidth}
    \centering
    \includegraphics[width=0.95\textwidth]{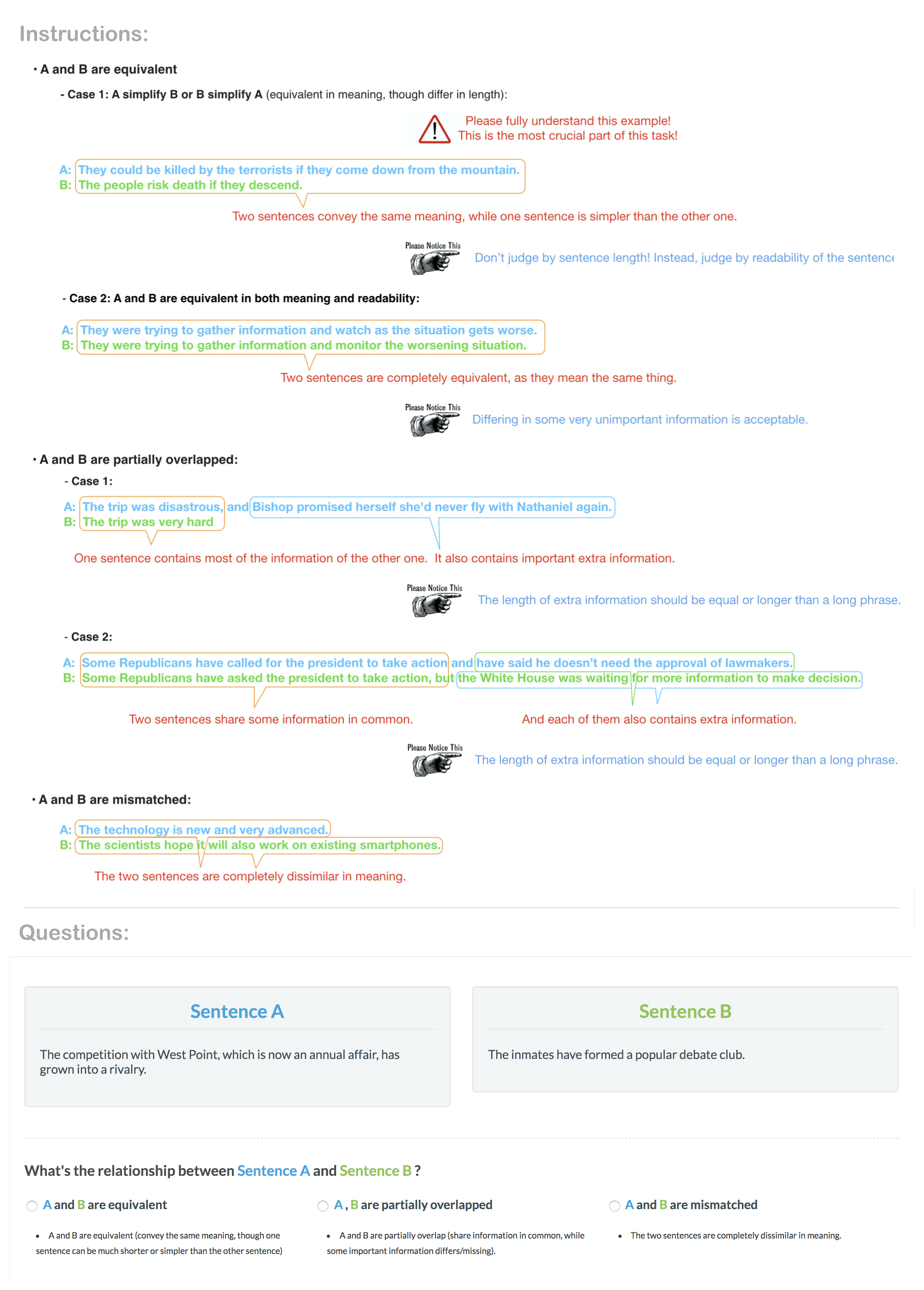}
    \captionof{figure}{Instructions and an example question for our crowdsourcing annotation on the Figure Eight platform.}
    \label{fig:annotation-}
\end{minipage}

\clearpage
\subsection{In-house Annotation Interface}
\label{appendix:human-annotation-inhourse-alignment}

\noindent\begin{minipage}{\textwidth}
    \centering
    \includegraphics[width=1\textwidth]{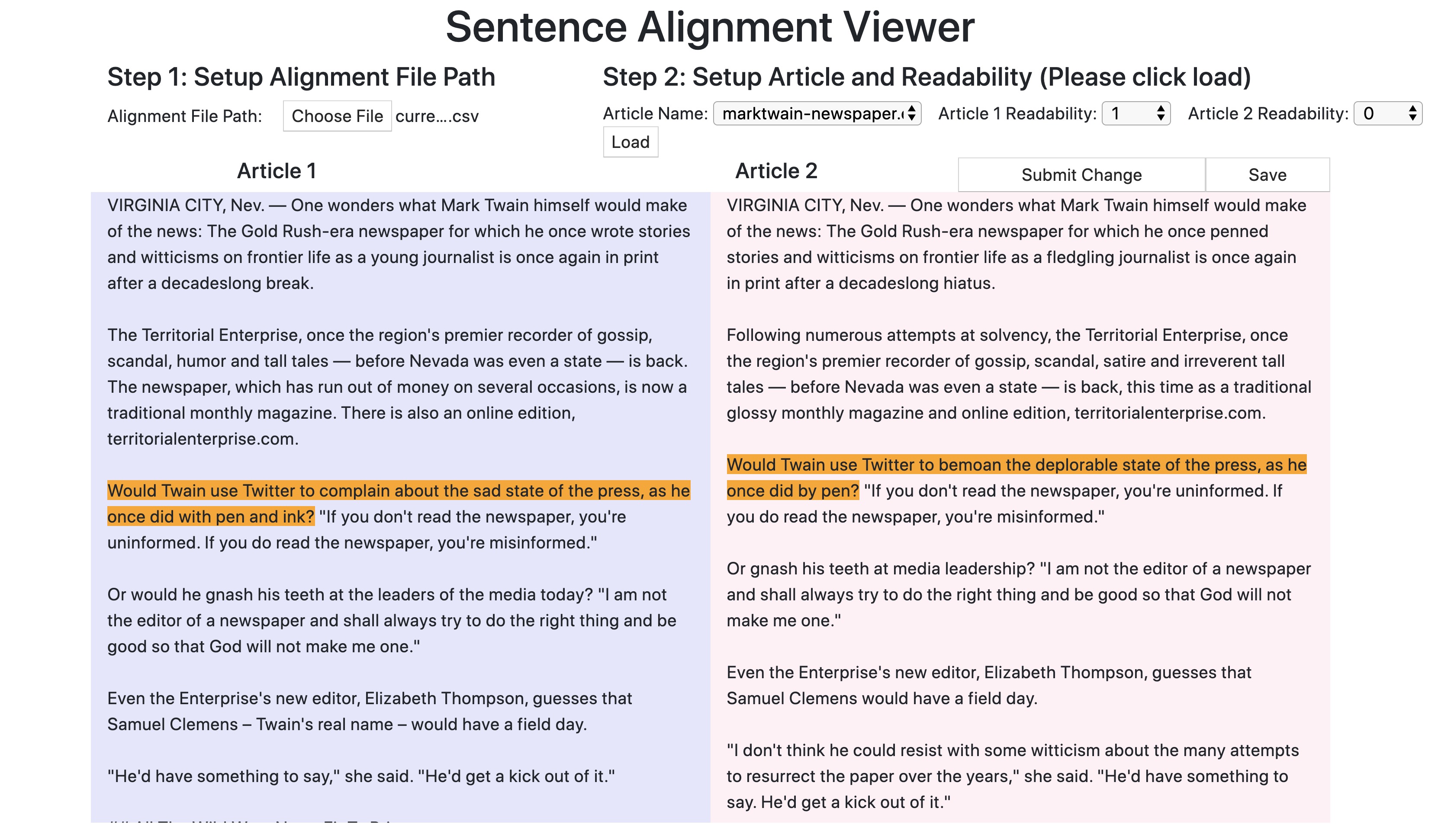}
    \captionof{figure}{Annotation interface for correcting the crowdsourced alignment labels.}
    \label{fig:annotation-}
\end{minipage}

\end{document}